\documentclass[conference,final]{IEEEtran}

\def\centerbmp#1#2#3{\vskip#2\relax\centerline{\hbox to#1{\special
  {bmp:#3 x=#1, y=#2}\hfil}}}

\def\centereps#1#2#3{\vskip#2\relax\centerline{\hbox to#1{\special
  {eps:#3 x=#1, y=#2}\hfil}}}

\usepackage{epsfig}
\usepackage{enumerate}
\usepackage{amsmath}
\usepackage{amsthm}
\usepackage{amscd}
\usepackage{amssymb}
\usepackage{rotating}
\usepackage{subfigure}
\usepackage{array}
\usepackage[noadjust]{cite}
\usepackage{caption2}
\usepackage{multirow}
\usepackage{bm}

\begin{document}

\bstctlcite{IEEEexample:BSTcontrol}
%

\title{Alleviating Overfitting for Polysemous Words for Word Representation Estimation Using Lexicons}

\author{\IEEEauthorblockN{Yuanzhi Ke}
\IEEEauthorblockA{Graduate School of Science and Technology\\
Keio University\\
Yokohama, Japan\\
Email: enshi@soft.ics.keio.ac.jp}
\and
\IEEEauthorblockN{Masafumi Hagiwara}
\IEEEauthorblockA{Graduate School of Science and Technology\\
Keio University\\
Yokohama, Japan\\
Email: hagiwara@soft.ics.keio.ac.jp
}}

\maketitle

\begin{abstract}
Though there are some works on improving distributed word representations using lexicons, the improper overfitting of the words that have multiple meanings is a remaining issue deteriorating the learning when lexicons are used, which needs to be solved. An alternative method is to allocate a vector per sense instead of a vector per word. However, the word representations estimated in the former way are not as easy to use as the latter one. Our previous work uses a probabilistic method to alleviate the overfitting, but it is not robust with a small corpus.
In this paper, we propose a new neural network to estimate distributed word representations using a lexicon and a corpus. We add a lexicon layer in the continuous bag-of-words model and a threshold node after the output of the lexicon layer. The threshold rejects the unreliable outputs of the lexicon layer that are less likely to be the same with their inputs. In this way, it alleviates the overfitting of the polysemous words. The proposed neural network can be trained using negative sampling, which maximizing the log probabilities of target words given the context words, by distinguishing the target words from random noises.
We compare the proposed neural network with the continuous bag-of-words model, the other works improving it, and the previous works estimating distributed word representations using both a lexicon and a corpus. The experimental results show that the proposed neural network is more efficient and balanced for both semantic tasks and syntactic tasks than the previous works, and robust to the size of the corpus.
\end{abstract}

\IEEEpeerreviewmaketitle

\section{Introduction}
\label{sec:intro}

Natural language processing is still a challenging research area of artificial intelligence. Especially, it is a difficult issue to recognize and represent the implicit features of a piece of text properly.

Distributed text representations estimated using a neural network are useful to be applied to conventional natural language processing algorithms \cite{bengio2003neural, mnih2007three, mnih2009scalable, collobert2011natural, huang2012improving, mikolov2012, mikolov2013distributed, DBLP:journals/corr/abs-1301-3781, mikolov2013linguistic, pennington2014glove, bojanowski2016enriching}. 
Great improvement by the distributed text representations estimated this way has been reported in name entity recognition and chunking~\cite{turian2010word}, text classification~\cite{socher2012semantic, le2014distributed, kim2014, joulin2016bag}, topic extraction~\cite{das-zaheer-dyer:2015, li2016generative}, and machine translation~\cite{DBLP:journals/corr/ZarembaSV14, DBLP:journals/corr/SutskeverVL14} etc.

However, some natural language processing tasks are still challenging. For example, the conventional algorithms fail to correctly predicate the number of starts of 40\% Amazon reviews~\cite{joulin2016bag}. It indicates the needs of higher quality text representation to improve the conventional algorithms.

The early approaches to estimate text representations use n-gram models~\cite{bengio2003neural, collobert2011natural, huang2012improving}. Mikolov et al. propose continuous bag-of-Awords and skip-gram models~\cite{mikolov2013distributed, DBLP:journals/corr/abs-1301-3781}. Their method outperforms the previous algorithms and costs less time. Pennington et al.~\cite{pennington2014glove} propose an algorithm using both local information and global information in the corpus and report a higher performance. Bojanowski et al.~\cite{bojanowski2016enriching} extend the models of Mikolov et al. using the character-level information of words. Their reported experimental results outperform the original models in syntactic tasks but fail to achieve an obvious improvement in semantic tasks. 

Lexicons are useful for us humans to learn a language. We can use them to help machines to learn natural languages as well. 
Chen et al.~\cite{chen2014unified} use the definitions in the lexicons to estimate representations for word senses and outperform the sense representations by Huang et al.~\cite{huang2012improving}. Other researches take advantage of the defined synonyms or paraphrases in lexicons. Yu et al.~\cite{yu2014improving} and Bollegala et al.~\cite{Bollegala2016} estimate the word representations by not only maximizing the probability of target word given a context, but also minimizing the distance of the paraphrases in a lexicon at the same time. Faruqui et al.~\cite{faruqui:2014:NIPS-DLRLW} propose a method refining trained word representation vectors using lexicons. Xu et al.~\cite{xu2014rc} estimate the word representations jointly by minimizing the distance of the tail word from the sum of the vectors of the head word and the relation for a triplet of words ($head, relation, tail$), and making words less similar to each other in a larger category. However, even though the previous methods using the paraphrases in lexicons reported improvements in syntactic analogical tasks, all of them failed to outperform the baselines in semantic analogical tasks.


However the previous works above using paraphrases in lexicons to improve estimated distributed word representations have not touched such an issue: For polysemous words that have different synonyms in different contexts, if we use the paraphrases to represent them without disambiguation, they may be over-fitted to improper senses. 

Some prior works try to solve similar problems by word sense disambiguation\cite{huang2012improving, chen2014unified}. However, word sense disambiguation is another difficult issue. Besides, it is less easy to use if a word has several vectors for its different senses instead of one vector per word, because the usage of such word representations in the conventional systems requires additional word sense disambiguation. Moreover, the word sense disambiguation in the usage needs to be the same as that for training, or the difference brought by different algorithms, granularity or hyperparameters may deteriorate the performance.

Another alternative approach~\cite{2016arXiv161100674K} considers the lexicon as a fuzzy set of paraphrases and using Bernoulli distribution subjected to the membership function of paraphrases to alleviate the problem. Although the method outperforms the previous works in a large corpus, the experimental results show that it is weak at small corpora.

In this paper, we propose a new method to improve the distributed word representations using paraphrases in a lexicon that is able to alleviate the overfitting of the polysemous words without disambiguation. Our method is efficient and easy to be combined with the conventional algorithms estimating word representations using corpora. In the experiment, our method outperforms the previous methods using paraphrases and keeps efficient for corpora in different sizes.

\section{Related Works}
\subsection{Continuous Bag-of-Words}

\begin{figure}[tb]
\begin{center}
\includegraphics[width=0.2\textwidth]{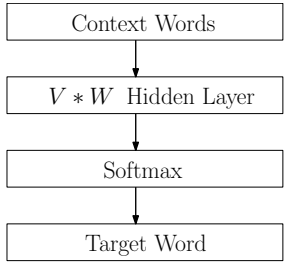}
\end{center}
\caption{Continuous Bag-of-Words (CBOW) Model.}
\label{fig_cbow}
\end{figure}

Continuous Bag-of-Words (CBOW) with negative sampling~\cite{DBLP:journals/corr/abs-1301-3781} is an efficient algorithm to estimate distributed word representations. The objective of CBOW is to maximize the log probability of a target word given the vectors of the context words. Denote the size of the vocabulary as $V$, the size of the word vectors as $W$. The model is like Fig. \ref{fig_cbow}.

Negative Sampling is an efficient method to maximize the log probability. It is a simplified Noise Contrastive Estimation~\cite{gutmann2012noise}. It trains the model by distinguishing target from randomly drawn noise. Denote the input vector of target word as $\bm{v}$, the ooutput vector as $\bm{v'}$, for each context word of the target word, the objective is to maximize:

\begin{equation}
\log \sigma ({v'_{w_O}}^{\rm T}v_{w_I}) + \sum_{i=1}^{n}E_{w_i} \sim P_n(w)[\log \sigma (-{v'_{w_i}}^{\rm T}v_{w_I})].
\end{equation}

Here, $P_n(w)$ is the distribution of the noise. $\sigma$ is a sigmoid function, $\sigma(x)=1/(1+e^{-x})$.

\subsection{Continuous Bag of Fuzzy Paraphrases}

\begin{figure}[tb]
\begin{center}
\includegraphics[width=0.32\textwidth]{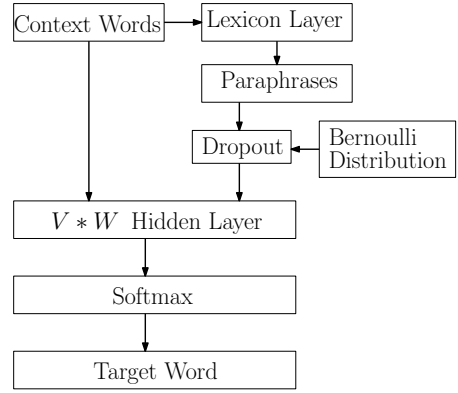}
\end{center}
\caption{Continuous Bag of Fuzzy Paraphrases (CBOFP).}
\label{fig_cbofp}
\end{figure}

Continuous Bag of Fuzzy Paraphrases (CBOFP)~\cite{2016arXiv161100674K} proposed by Ke et al. is a model based on CBOW to learn word representations using both a lexicon and a corpus. It is able to alleviate the overfitting of the word vectors for polysemous words. It outperforms the previous works using lexicons to estimate distributed word representations, but is not robust for small corpora. Fig. \ref{fig_cbofp} shows the structure of the model.

CBOFP adds a lexicon layer to CBOW. Unlike the previous work using lexicons to estimate distributed word representations, in CBOFP, every paraphrase is a fuzzy member of the paraphrase set of a word with a degree of truth. The outputs of the lexicon layer are dropped out randomly. The dropout is controlled by a function of the paraphrases' degrees of truth that returns 0 or 1 drawn from a Bernoulli distribution. Denote the degree of truth as $x$, the control function $f(x)$ is defined as the following:

\begin{equation}
f(x) \sim Bernoulli(x).
\end{equation}

The degree of truth is measured using the score provided by a paraphrase database called PPDB~\cite{ganitkevitch2013ppdb, PavlickEtAl-2015:ACL:PPDB2.0, PavlickEtAl-2015:ACL:Semantics}. Denote the score as $S$, the paraphrases set as $L$, the degree $x$ is calculated as the following:

\begin{equation}
x={\frac{S}{\max \limits_{L}S}}.
\end{equation}

The reported experimental results show that this method outperforms the previous works, especially in semantic tasks, but not robust with small corpora such as text8\footnote{http://mattmahoney.net/dc/text8.zip}.

\section{The proposed method}
\label{sec:met}
\subsection{Structure}

Our previous work CBOFP is weak at small corpus because it involves a probabilistic method to alleviate the overfitting of polysemous words. Such a method requires enough amount of training data. Thus we consider another method not using a probabilistic way. 

\begin{figure}[tb]
\begin{center}
\includegraphics[width=0.45\textwidth]{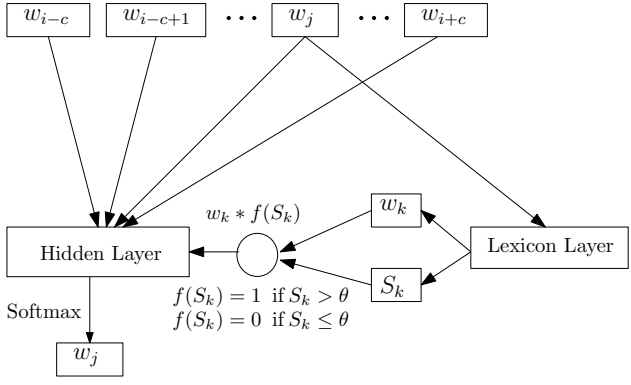}
\end{center}
\caption{The proposed neural network. For each input of the lexicon layer, it outputs a paraphrase $w_k$ and a score $S_k$. For every output of the lexicon layer, we compare $S_k$ with predefined threshold $\theta$. The outputs whose scores are greater than $\theta$ will attend the training of word vectors in the hidden layer. Otherwise, the outputs whose scores are less than $\theta$ are dropped out.}
\label{fig_thr}
\end{figure}

Instead of dropping out some of the outputs of the lexicon layer randomly, we add a node after the lexicon layer as shown in Fig. \ref{fig_thr}. The inputs are the context word of the target word. They are both inputs into the hidden layer that contains the word vectors to learn, and input into the lexicon layer. The lexicon layer outputs the paraphrases of the inputs and their scores. The node after the lexicon layer takes the score of the paraphrase and holds a threshold. If the score of the paraphrase is higher than the threshold, it returns true. Otherwise, it returns false. The paraphrases whose outputs are false are not learned. True paraphrases of the context words are input into the hidden layer to learn together with the original context words. The output of the hidden layer is a vector of the vocabulary size. After a softmax function, the neural network outputs the target word. By maximizing the probability of outputting correct word, the hidden layer can be trained and used as the vectors represent the words in the vocabulary.

\subsection{The Lexicon Layer}
\label{lex_layer}

We use a paraphrase database called PPDB2.0~\cite{ganitkevitch2013ppdb, PavlickEtAl-2015:ACL:PPDB2.0, PavlickEtAl-2015:ACL:Semantics} to build our lexicon layer, which has been used in the previous works~\cite{2016arXiv161100674K, yu2014improving, faruqui:2014:NIPS-DLRLW}. The paraphrases in it are extracted automatically from multilingual resources. It is reported useful in many other tasks such as recognizing textual entailment~\cite{beltagy2014utexas, bjerva2014meaning}, measuring the semantic similarity~\cite{han2013umbc, ji2013discriminative, sultan2014dls}, monolingual alignment~\cite{sultan2014back, yao2013semi}, and natural language generation~\cite{ganitkevitch2011learning}.

PPDB2.0 provides not only the paraphrases but also the features, the alignment types and the entailment types. There are six types of entailment in PPDB2.0 as shown in Table \ref{para_type}. 

We compared the performance of our models that employed different paraphrases. The experiments are described and discussed in Subsection \ref{differ_para}.

\begin{table}[b]
\caption{Different types of relationships of paraphrases in PPDB2.0~\cite{PavlickEtAl-2015:ACL:PPDB2.0, PavlickEtAl-2015:ACL:Semantics}.}
\label{para_type}
\renewcommand\arraystretch{1.5}
\begin{center}
    \begin{tabular}{|l|l|}
    \hline
    Relationship  			& Description   \\ \hline
    Equivalence  			& $X$ is the same as $Y$ \\ \hline
    Forward Entailment 	& $X$ is more specific than/is a type of $Y$   \\ \hline
    Reverse Entailment 	& $X$ is more general than/encompasses $Y$   \\ \hline
    Exclusion	& $X$ is the opposite of $Y$ / $X$ is mutually exclusive with $Y$ \\ \hline
    Other			& $X$ is related in some other way to $Y$ \\ \hline
    Independent     		& $X$ is not related to $Y$   \\ \hline
    \end{tabular}
\end{center}
\end{table}

As the score input to threshold node, we use the PPDB2.0 scores. They are estimated by a supervised scoring model on the basis of human judgments for 26,455 paraphrase pairs and high correlation of the PPDB2.0 scores and 
 human judgments are reported~\cite{PavlickEtAl-2015:ACL:PPDB2.0}.

\subsection{Learning the Word Representations}

With the threshold, the objective is to maximize

\begin{equation}
\label{eq_lp}
\sum_{w_i \in G}^{G}\sum_{j \in C}\left [ \log p(w_i|w_j)+\sum_{w_k \in L_{w_j}}^{L_{w_j}} f(S_{jk})\log p(w_i | w_k) \right ].
\end{equation}

Here, $G$ is the set of words in the corpus, $C$ is the context of word $w_i$, $L_{w_j}$ is the paraphrase set of the context word $w_j$, $S_{jk}$ is the score of paraphrase $w_k$ in $L_{w_j}$. The function $f(S_{jk})$ is defined as the following: 

\begin{equation}
f(S_{jk})\begin{cases}
1 & \text{ if } S_{jk}>\theta  \\ 
0 & \text{ if } S_{jk}<\theta .
\end{cases}
\end{equation}

Here, $\theta$ is the threshold of the threshold node.

The log probability is maximized in the learning phrase using negative sampling. Similarly to that for CBOW, we maximize the log probability in equation (\ref{eq_lp}) by maximizing

\begin{equation}
\begin{split}
\log \sigma ({v_{w_i}}^{\rm T}v_{w_j}) + \sum_{n=1}^{N}E_{w_i} \sim P_n(w)[\log \sigma (-{v_{w_n}}^{\rm T}v_{w_j})], \\
w_n \neq w_i, w_n \notin L_{w_j}.
\end{split}
\end{equation}

We draw noise that does not equal the target word or is not in the paraphrase set of the target word from the noise distribution $P_n(w)$. The target of the noise is labeled with zero, and the neural network is trained by maximizing the probability of the target word, given the input from the input layer and the lexicon layer, while minimizing the probability of the target word, given the noise, at the same time.

\section{Experiments}

\subsection{The Corpus Used in the Experiments}

We use text8 and a larger corpus called enwiki9\footnote{http://mattmahoney.net/dc/enwiki9.zip} for the experiments because we are to evaluate if the proposed model is more robust for a smaller corpus than the previous work and keeps efficient for a larger one. Both text8 and enwiki9 are part of the English Wikipedia\footnote{https://en.wikipedia.org/} dump. Text8 contains 16,718,843 tokens while enwiki9 contains 123,353,508 tokens. The vocabulary size of text8 is 71,291, while that of enwiki9 is 218,317. We see that text8 is one tenth the size of enwiki9.

\subsection{The Task for Evaluation}

The word analogical reasoning task introduced by Mikolov et al.\cite{DBLP:journals/corr/abs-1301-3781} is used for evaluation in the experiments. For a quaternion of words ($w_A$, $w_B$, $w_C$, $w_D$) in which the relationship of $w_A$ and $w_B$ is similar to that of $w_C$ and $w_D$, the objective is to predict $w_D$ on the basis of $w_A$, $w_B$ and $w_C$ by searching the word whose vector is the closest to $v_B-v_A+v_C$. The dataset has a semantic part and a syntactic part. In the semantic part, ($w_A$, $w_B$) and ($w_C$, $w_D$) have a similar semantic relationship while they have a syntactic one in the syntactic part.
Table \ref{wart} shows an example of the questions in the task.

There are 8,869 questions in the semantic part and 10,675 questions in the syntactic part.

\begin{table}[tb]
\caption{Examples of the questions in the word analogical reasoning task.}
\label{wart}
\renewcommand\arraystretch{1.5}
\begin{center}
\begin{tabular}{|c|c c c c|}
\hline
\multirow{2}{*}{Semantic}  & Beijing & China     & Tokyo   & Japan   \\ \cline{2-5} 
                           & boy     & girl      & brother & sister  \\ \hline
\multirow{2}{*}{Synatctic} & amazing & amazingly & calm    & calmly  \\ \cline{2-5} 
                           & aware   & unaware   & clear   & unclear \\ \hline
\end{tabular}
\end{center}
\end{table}

\subsection{Comparison of Different Paraphrase Types}
\label{differ_para}

As described in Subsection \ref{lex_layer}, there are six different paraphrase types in PPDB2.0. We used paraphrases of different types to train 100-dimensional word vectors in the hidden layer with text8. The context window was eight. The threshold $\theta$ is set to $3.5$. Then we used the trained word vectors to do the word analogical reasoning task. The results are shown in Table \ref{tab:para}. 

We also compare the cosine similarities of words with those in the gold standard dataset Simlex-999~\cite{hill2016simlex}. Simlex-999 contains 999 word pairs. Each pair is annotated with a semantic similarity by humans. The annotators were told to be careful about the difference of ``semantically similarity" and ``relatedness." They give high scores to ``semantical similar words" like ``coast" and ``shore" but give low scores to ``related but not similar words" like ``clothes" and ``closet." We use it to see how the word vectors capture the meaning of words. We report the Spearman's rank correlations $\rho$. The results are shown in Table \ref{tab:para2}.

\begin{table}[bp]
	\centering
	\caption{Correct answer rates of word analogical reasoning tasks when paraphrases of different types were used. ``Entailment" set in every test includes ``forward entailment" and ``reverse entailment." ``Sem" refers to the semantic part. ``Syn" refers to the syntactic part.}
	\label{tab:para}%
	\renewcommand\arraystretch{1.5}
	\begin{tabular}{|l|c|c|c|}
		\hline
		Paraphrase Set& Sem [\%] & Syn [\%] & Total [\%] \\ \hline
		Equivalence & 44.46  & 38.88  & 41.20  \\ \hline
		Entailment (Forward+Reverse) & 43.57  & 39.29  & 41.07  \\ \hline
		Equivalence+Entailment & 44.93  & 39.31  & 41.65  \\ \hline
		Equivalence+Entailment+Other & 44.89  & 39.82  & \textbf{41.93}  \\ \hline	
		Equivalence+Entailment+Other+Exclusive & \textbf{44.96}  & 38.81  & 41.37  \\ \hline
		Equivalence+Entailment+Other+Independent & 44.19  & \textbf{40.22}  & 41.87  \\ \hline
	\end{tabular}%
\end{table}%

\begin{table}[htbp]
	\centering
	\caption{Spearman's rank correlations $\rho$ with Simlex-999~\cite{hill2016simlex} when paraphrases of different types were used. ``Entailment" set in every test includes ``forward entailment" and ``reverse entailment."}
	\label{tab:para2}%
	\renewcommand\arraystretch{1.5}
	\begin{tabular}{|l|c|}
		\hline
		Paraphrase Set& $\rho$ \\ \hline
		Equivalence & 0.2883  \\ \hline
		Entailment(Forward+Reverse) & \textbf{0.2966}  \\ \hline
		Equivalence+Entailment & 0.2920  \\ \hline
		Equivalence+Entailment+OtherRelated & 0.2915  \\ \hline
		Equivalence+Entailment+OtherRelated+Exclusive & 0.2874  \\ \hline
		Equivalence+Entailment+OtherRelated+Independent & 0.2913  \\ \hline
	\end{tabular}%
\end{table}%

In Table \ref{tab:para}, we can see that using ``equivalence" and ``entailment" together is better than using either of them alone. Employing paraphrases of ``other" type together with ``equivalence" and ``entailment" achieves better performance in syntactic tests, best total correct answer rates but is weaker in semantic tests. Adding paraphrases of ``exclusive" type improves the correct answer rate of semantic tests but deteriorate that of syntactic tests. Involving paraphrases of ``independent" type improves the correct answer rate of syntactic tests but deteriorate that of semantic tests.

In Table \ref{tab:para2}, we find that use paraphrases of ``forward entailment" and ``reverse entailment" makes the cosine similarities of words more close to the human annotated similarities in Simlex-999. Using the other types deteriorates the scores.

We can see that what is the best paraphrase set dependents on the requirement of the task. The word analogical reasoning task contains lots of questions about ``topical relatedness" such as ``Athens - Greece = Beijing - China." However, Simlex-999 does not consider such related words are similar. Similarly, many paraphrases of ``other" and ``independent" are topical related or unrelated. Therefore we can see improvement for word analogical reasoning tasks but deterioration for Simlex-999 when we employed such paraphrases. Besides, different paraphrase sets hold the different balance of ``similarity" and ``relatedness." It is the reason of the differences of their performances for different benchmarks.

\subsection{Parameter Tuning}

\begin{figure*}[htbp]
\begin{center}
\includegraphics[width=0.8\textwidth]{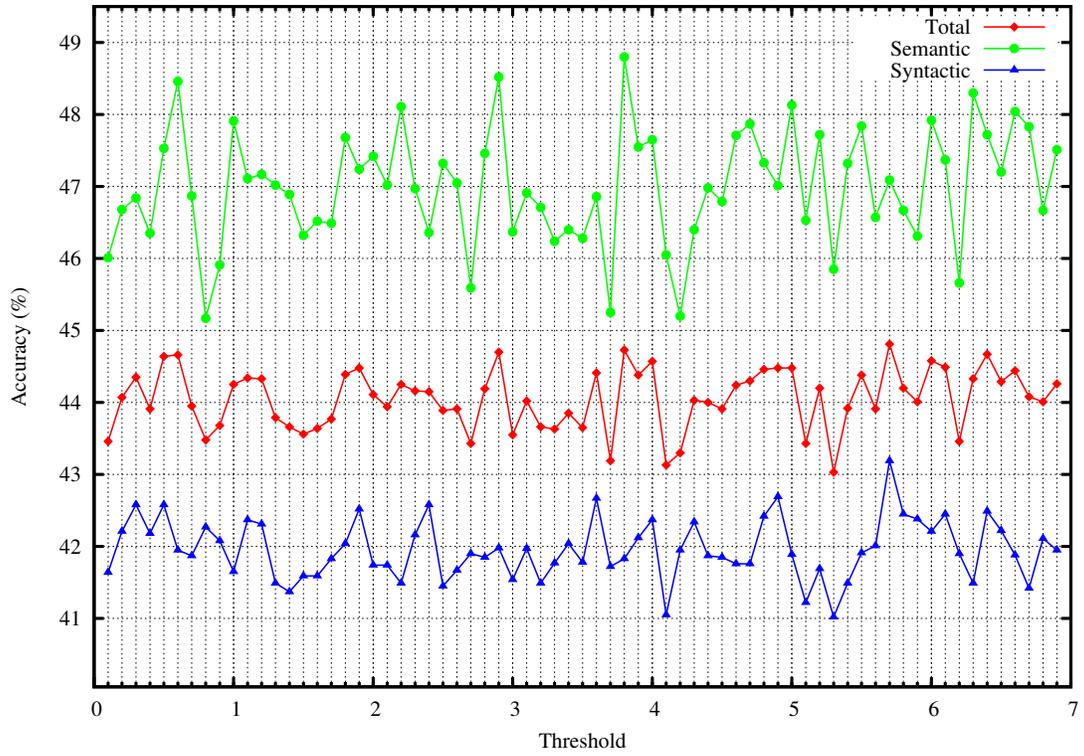}
\end{center}
\caption{Threshold tuning for text8.}
\label{fig_tunt8}
\end{figure*}

\begin{figure*}[htbp]
\begin{center}
\includegraphics[width=0.8\textwidth]{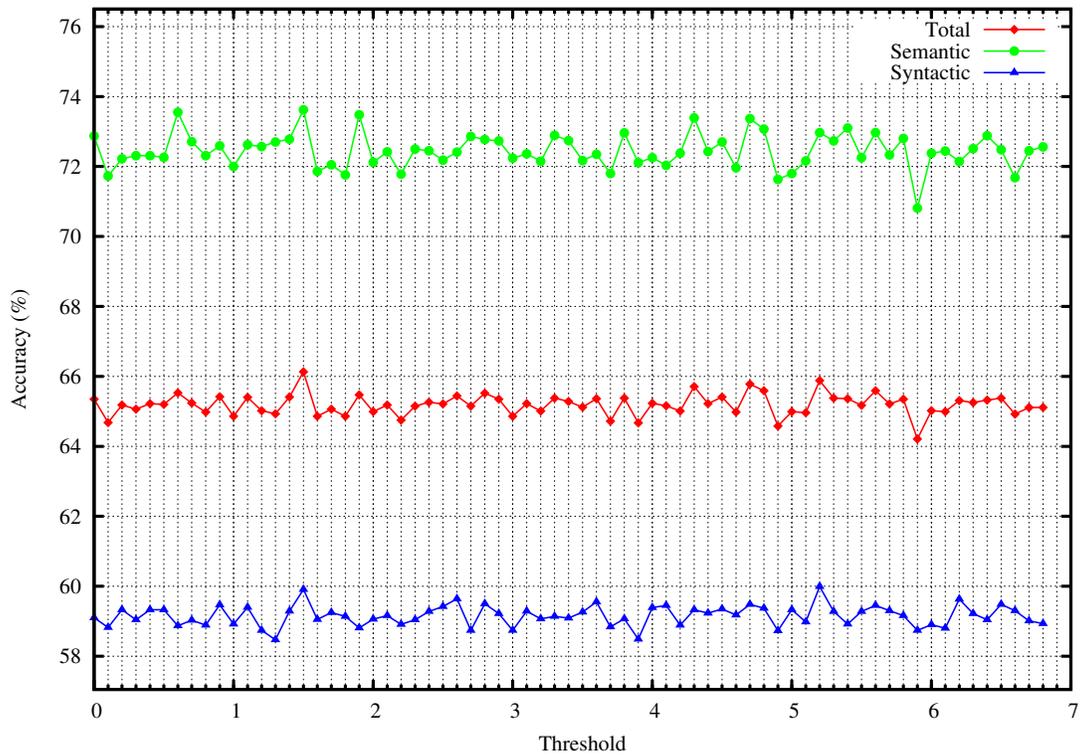}
\end{center}
\caption{Threshold tuning for enwiki9.}
\label{fig_tunt9}
\end{figure*}

To find the proper value of threshold $\theta$, we let the proposed neural network learn the word representations for text8 and enwiki9 and run the word analogical reasoning task with different thresholds. The paraphrases of ``equivalence", ``forward entailment" and ``reverse entailment" are used (see Table \ref{para_type}). The results for text8 is shown in Fig. \ref{fig_tunt8}. The results for enwiki9 is shown in Fig. \ref{fig_tunt9}.

We see that the correlation of the performance and the threshold is not linear. And it is not very same for different tasks or corpora. However, because all of the score of the paraphrases \textemdash the PPDB2.0 scores are less than seven in the version we used, we can find the best threshold in the interval.

From Fig. \ref{fig_tunt8}, we see that the best threshold for the semantic part and the total dataset is 3.8, using text8. The best threshold for the syntactic part is 5.7. From Fig. \ref{fig_tunt9}, we see that the best threshold for the semantic part and the total dataset is 1.5, using enwiki9. The best threshold for the syntactic part is 5.2. It shows that lower threshold is better for a larger corpus and semantic tasks. The best correct answer rates are shown in Table \ref{best}.

The other parameters are set as those used by the public word2vec demo\footnote{https://code.google.com/archive/p/word2vec/} that are already well tuned. The initial learning rate is set to 0.05. The number of negative samples drawn in negative sampling is set to 25. The context window is set to 8. The total iteration time is set to 25. And the size of the word vectors in the hidden layer is set to 200.

\subsection{Comparison with Other Methods}


\begin{table}[tb]
\caption{The best correct answer rates in the word analogical reasoning task.}
\label{best}
\renewcommand\arraystretch{1.5}
\begin{center}
    \begin{tabular}{|c|c|c|c|}
    \hline
    ~         & Semantic [\%]   & Syntactic [\%]  & Total[\%] \\ \hline
    text8  & 48.80 & 47.09 & 44.73 \\ \hline
    enwiki9 & 73.62 & 59.63 & 66.13  \\ \hline
    \end{tabular}
\end{center}
\end{table}

We compare our proposed neural network with CBOW~\cite{DBLP:journals/corr/abs-1301-3781}, CBOW enriched with subword information~\cite{bojanowski2016enriching}, GloVe~\cite{pennington2014glove}, the work of Faruqui et al.~\cite{faruqui:2014:NIPS-DLRLW}, jointReps~\cite{Bollegala2016}, RC-Net~\cite{xu2014rc} and CBOFP~\cite{2016arXiv161100674K}. For the first four, we got their public online available implements by the authors. We used the implements by the authors to learn word representations using text8 and enwiki9 and report the results. For jointReps and RC-Net, we failed to find an available implements that correctly run using text8 and enwiki9. Thus we use the results in their papers for jointReps and RC-Net. The reported results of RC-Net are achieved using enwiki9 as well, while those of jointReps are achieved using ukWaC\footnote{http://wacky.sslmit.unibo.it}.

In Table \ref{compare_t8}, we compare our proposed neural network under text8 with CBOW, CBOW enriched with subword information, GloVe, the work of Faruqui et al. and CBOFP. We do not compare with jointReps and RC-Net here, because there are no reported results evaluated by the word analogical reasoning tasks of these methods with a similar corpus. 

We can see that using text8, our proposed neural network achieves the best accuracy in the semantic, and outperforms the others except CBOW enriched with subword information in syntactic part. CBOW enriched with subword information is reported powerful for representing syntactic features, but ours is more balanced.

In Table \ref{compare_t9}, we compare our proposed neural network with the previous works using enwiki9. The results of jointReps and RC-Net here are the reported results in their paper. The results of JointReps are not achieved using enwiki9 but using ukWaC.

We see that our proposed neural network outperforms the previous works for the semantic part and the whole dataset under enwiki9. The proposed neural network failed to outperform CBOW enriched with subword information in the syntactic part. However, our proposed neural network is better at the semantic part than the CBOW enriched with subword information and outperforms it in the overall accuracy.

All of the experimental results show that the proposed neural network is more balanced than the previous works, more powerful than the other works using lexicons to estimate or improve distributed word representations, benefiting from the threshold node alleviating the overfitting of polysemous words.
Moreover, while the CBOFP failed to outperform most of the others under text8, the proposed neural network keeps outperform all the others in semantic parts and achieves the second best overall accuracy. It shows that the proposed neural network is more robust to the size of the corpus.

\begin{table}[tb]
\caption{Comparison against the previous works using text8.}
\label{compare_t8}
\renewcommand\arraystretch{1.5}
\begin{center}
    \begin{tabular}{|c|c|c|c|}
    \hline
    ~         & Semantic [\%]   & Syntactic [\%]  & Total[\%] \\ \hline
    \textbf{The proposed}	& \textbf{48.80} & 47.09 & 44.73   \\ \hline
    CBOW~\cite{DBLP:journals/corr/abs-1301-3781} & 46.72 & 41.90 & 43.91 \\ \hline
    Enriched CBOW\cite{bojanowski2016enriching}  & 15.95 & \textbf{73.62} & \textbf{49.63}  \\ \hline
    GloVe~\cite{pennington2014glove}     		& 41.15 & 21.59 & 29.73  \\ \hline
    Faruqui~\cite{faruqui:2014:NIPS-DLRLW}   	& 34.80 & 51.53 & 44.57    \\ \hline  
    CBOFP~\cite{2016arXiv161100674K}  			& 46.35 & 42.13 & 43.88 \\ \hline
    \end{tabular}
\end{center}
\end{table}

\begin{table}[tb]
\caption{Comparison against the previous works using enwiki9.}
\label{compare_t9}
\renewcommand\arraystretch{1.5}
\begin{center}
    \begin{tabular}{|c|c|c|c|}
    \hline
    ~         & Semantic [\%]   & Syntactic [\%]  & Total[\%] \\ \hline
    \textbf{The proposed}	& \textbf{73.62} & 59.63 & \textbf{66.13}   \\ \hline
    CBOW~\cite{DBLP:journals/corr/abs-1301-3781} 				& 72.65 & 59.25 & 65.33  \\ \hline
    Enriched CBOW\cite{bojanowski2016enriching}  & 33.08 & \textbf{75.39} & 56.19  \\ \hline
    GloVe~\cite{pennington2014glove}     		& 66.35 & 43.46 & 53.80  \\ \hline
    Faruqui~\cite{faruqui:2014:NIPS-DLRLW}   		& 53.88 & 61.31 & 57.94    \\ \hline  
    JointReps~\cite{Bollegala2016} 		& 61.46		& 69.33  & 65.76  \\ \hline
    RC-Net~\cite{xu2014rc}     		& 34.36	& 44.42	& -   \\ \hline
    CBOFP~\cite{2016arXiv161100674K}  			& 73.29 & 59.44 & 65.85 \\ \hline
    \end{tabular}
\end{center}
\end{table}

\section{Conclusions}
\label{sec:con}

To alleviate the overfitting of the polysemous words, we proposed a new neural network estimating distributed word representations in this paper. Additional to the conventional continuous bag-of-words model, we added a lexicon layer, and a threshold node after the output of the lexicon layer. The threshold is manually tuned. The neural network can be trained using negative sampling. The experimental results show that the proposed neural network is more powerful and balanced than the previous models using lexicons to estimate or improve distributed word representations. Besides, unlike our previous work CBOFP, the proposed neural network in this paper is robust to small corpora. 
Automatic tunning of the threshold and the other parameters is a remaining issue. We are going to work on it in the future.

\bibliographystyle{IEEEtran}
\bibliography{ijcnn2017}

\end{document}